\documentclass[11pt, a4paper]{article}

\usepackage[utf8]{inputenc}
\usepackage[T1]{fontenc}
\usepackage{graphicx}
\usepackage{amsmath, amssymb}
\usepackage{booktabs} 
\usepackage{multirow}
\usepackage{longtable} 
\usepackage{url}
\usepackage{textgreek}
\usepackage{array} 
\usepackage{booktabs}
\usepackage{paracol}
\usepackage{colortbl} 
\usepackage{pifont}   
\usepackage[margin=1in]{geometry}
\usepackage[colorlinks=true, linkcolor=blue, citecolor=blue, urlcolor=blue]{hyperref}
\usepackage[super,sort&compress]{natbib} 
\usepackage{float}    
\usepackage{dirtree} 
\usepackage{xcolor}  
\usepackage{caption}
\newcommand{\cmark}{\textcolor{green!70!black}{\ding{51}}} 
\newcommand{\xmark}{\textcolor{red}{\ding{55}}}
\title{S1-MMAlign: A Large-Scale, Multi-Disciplinary Dataset for Scientific Figure–Text Understanding}

\author{
    He Wang$^{1,2,3,*}$, Longteng Guo$^{1,3,*}$, Pengkang Huo$^{1,3}$, Xuanxu Lin$^{1,3}$, \\
    \vspace{1mm}
    Yichen Yuan$^{1,3}$, Jie Jiang$^{1,3}$ \& Jing Liu$^{1,3,\dagger}$ \\
    \\
    \small $^{1}$ Institute of Automation, Chinese Academy of Sciences, Beijing, China \\
    \small $^{2}$ School of Advanced Interdisciplinary Sciences, University of Chinese Academy of Sciences, Beijing, China \\
    \small $^{3}$ University of Chinese Academy of Sciences, Beijing, China \\
    \\
    \footnotesize $^{*}$ These authors contributed equally to this work. \\
    \footnotesize $^{\dagger}$ Corresponding author: jliu@nlpr.ia.ac.cn
}
\date{}

\begin{document}

\maketitle

\begin{abstract}
Multimodal learning has revolutionized general domain tasks, yet its application in scientific discovery is hindered by the profound semantic gap between complex scientific imagery and sparse textual descriptions. We present S1-MMAlign, a large-scale, multi-disciplinary multimodal dataset comprising over 15.5 million high-quality image-text pairs derived from 2.5 million open-access scientific papers. Spanning disciplines from physics and biology to engineering, the dataset captures diverse visual modalities including experimental setups, heatmaps, and microscopic imagery. To address the pervasive issue of weak alignment in raw scientific captions, we introduce an AI-ready semantic enhancement pipeline that leverages advanced multimodal large language models to recaption images, by synthesizing comprehensive context from paper abstracts and the citation contexts of corresponding figures. Technical validation confirms that our enhancement pipeline markedly improves data quality via reduced SciBERT pseudo-perplexity and enhanced CLIP image-text alignment, while also significantly boosting multimodal large language models’ performance in zero-shot scientific captioning, multi-domain scientific reasoning, and visual instruction tuning. S1-MMAlign provides a pivotal foundational resource for cross-modal scientific understanding in the AI for Science era, supporting the development of scientific foundation models and a wide range of downstream scientific intelligence applications.
\end{abstract}

\section*{Background \& Summary}

The paradigm of "AI for Science" is accelerating, shifting towards a data-driven era where scientific foundation models are poised to automate discovery and reasoning \cite{jumper2021highly, taylor2022galactica}. However, a critical bottleneck impedes the development of multimodal scientific agents: the profound \textbf{semantic misalignment} inherent in raw scientific publications. Unlike general domain images (e.g., COCO \cite{lin2014microsoft}, LAION \cite{schuhmann2022laion}) where captions offer self-contained visual descriptions (e.g., ``a cat sitting on a mat''), scientific figures are fundamentally distinct. They encapsulate complex logic, physical mechanisms, and variable relationships that are often implicitly defined \cite{hsu2021scicap}.

Critically, the textual captions found in raw papers are frequently \textbf{context-dependent and sparse}. A typical caption might read ``Figure 3: Ablation study results,'' relying entirely on the main text for interpretation. This indexical nature creates a significant semantic gap for machine learning models, as the visual signal is divorced from its theoretical grounding. Training on such data leads to superficial alignment, where models learn to recognize chart types but fail to comprehend the underlying scientific implications \cite{lin2023sciscinet}. Existing datasets often lack the structural depth to bridge this gap, serving merely as aggregations of raw, noisy pairs \cite{methani2020plotqa, taylor2022galactica}.

Despite recent advancements, existing scientific figure datasets remain severely limited by these semantic and structural gaps. As detailed in \textbf{Table \ref{tab:comparison}}, earlier efforts often restrict their scope to specific domains (e.g., Computer Science in SciCap \cite{hsu2021scicap} and M-Paper \cite{lin2023sciscinet}) or focus predominantly on single modalities like simple charts. Even large-scale, multi-disciplinary collections (such as MMArxiv \cite{del2024mmarxiv}) overwhelmingly rely on raw, uncurated captions that fail to capture the complex visual reasoning required for rigorous scientific analysis. 

\begin{figure*}[t] 
    \centering
    \includegraphics[width=\textwidth]{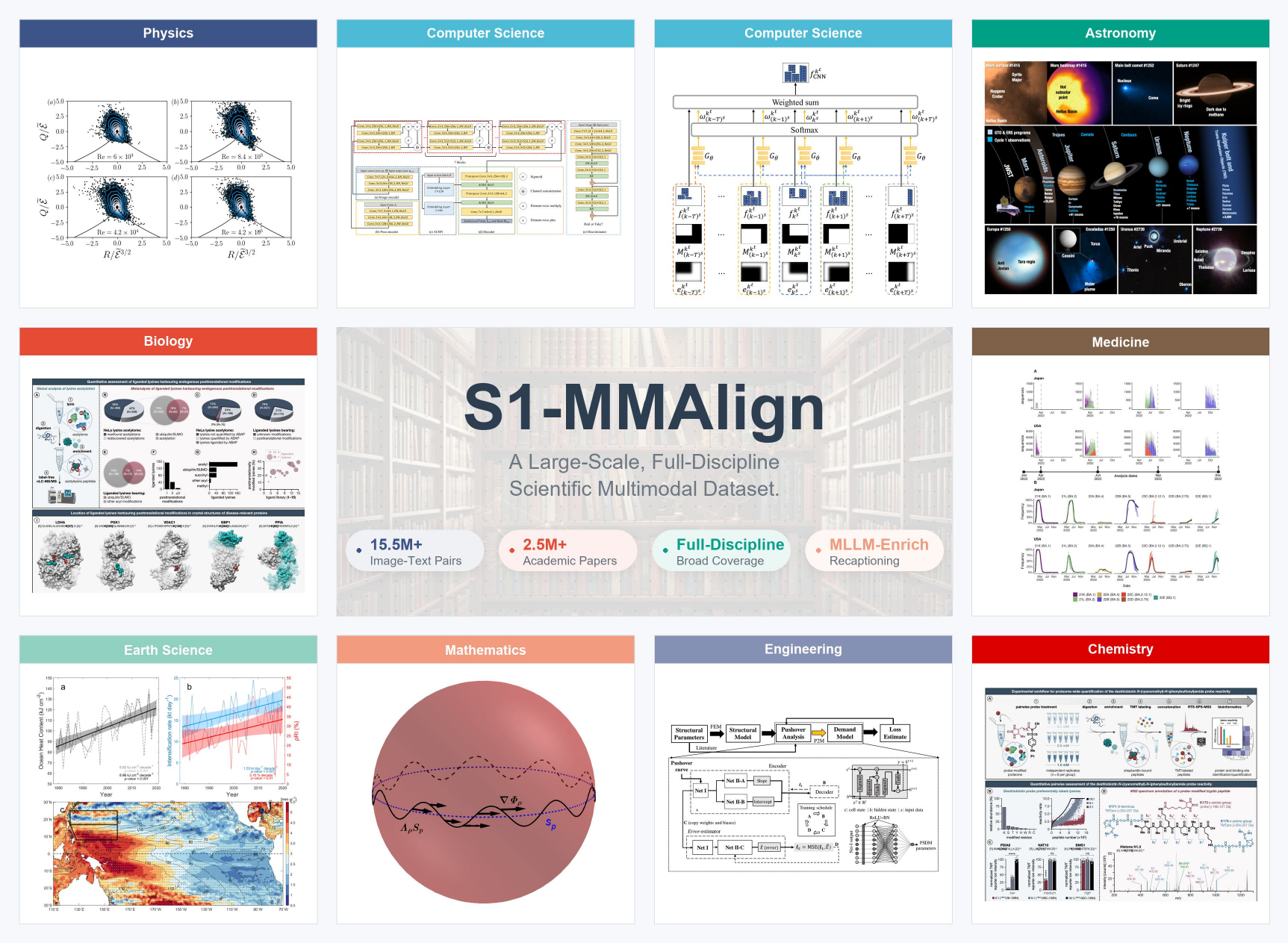} 
    \caption{
        \textbf{Visual diversity of the S1-MMAlign.} 
        A curated selection of sample figures demonstrating the broad range of visual modalities (e.g., line plots, microscopy images, block diagrams, and heatmaps) across various scientific disciplines covered in our collection.
    }
    \label{fig:teaser}
\end{figure*}

To directly overcome these deficiencies, we release \textbf{S1-MMAlign}, one of the largest full-discipline scientific multimodal data resources to date. Built from 2.5 million open-access academic papers, the dataset contains over \textbf{15.5 million} high-quality native figure-text pairs, marking a notable leap in scale and data granularity. As illustrated by the diverse samples in Figure \ref{fig:teaser}, S1-MMAlign encompasses a wide array of visual modalities across various scientific domains. Our core contribution is a novel context-aware figure-text alignment enhancement framework that fuses visual signals with paper-level context (e.g., abstracts, citation contexts). By combining full-discipline breadth, unprecedented scale, and MLLM-enriched recaptioning, S1-MMAlign mitigates the weak alignment of raw scientific captions, transforms sparse figures into self-contained knowledge units, and enables models to learn both the visual features and the underlying scientific significance of the content.

\begin{table}[H]
\centering
\renewcommand{\arraystretch}{1.2}
\resizebox{\textwidth}{!}{%
\begin{tabular}{@{} l c c r c c c @{}}
\toprule
\textbf{Dataset} & \textbf{Source Type} & \textbf{Subject} & \textbf{Figures} & \textbf{Modality} & \textbf{Context} & \textbf{Recaption} \\
\midrule
MMArxiv \cite{del2024mmarxiv} & arXiv & Multi. & 6.4M & Diverse & \cmark & \xmark \\
FigureQA \cite{kahou2017figureqa} & Synthetic & N/A & 140K & Charts & \xmark & \xmark \\
MMSCI \cite{he2024mmsci} & Nat. Commun. & Natural Sci. & 742K & Diverse & \cmark & \xmark \\
SciFiBench \cite{roberts2024scifibench} & arXiv & CS & 6K & Charts & \xmark & \xmark \\
SciCap \cite{hsu2021scicap} & arXiv & CS & 414K & Charts & \cmark & \xmark \\
M-Paper \cite{lin2023sciscinet} & arXiv & CS & 350K & Diverse & \cmark & \xmark \\
CharXiv \cite{wang2024charxiv} & arXiv & Multi. & 2.3K & Charts & \xmark & \xmark \\
DvQA \cite{kafle2018dvqa} & Synthetic & N/A & 300K & Charts & \xmark & \xmark \\

\midrule
\rowcolor[gray]{0.95} 
\textbf{S1-MMAlign} & \textbf{Multiple OA} & \textbf{Full-discipline} & \textbf{15.5M} & \textbf{Diverse} & \cmark & {\textbf{MLLM-Enriched}} \\
\bottomrule
\end{tabular}%
}
\caption{Comparison with previous scientific figure datasets. S1-MMAlign covers a broader range of scientific disciplines and offers the largest scale of figure-text pairs. Crucially, it provides context-aware enhanced captions, distinguishing it from previous datasets that rely solely on raw, sparse text. Diverse figures include charts, diagrams, microscopy images, molecular structures, and equation snippets.}
\label{tab:comparison}
\end{table}

\begin{table}[H]
\centering
\renewcommand{\arraystretch}{1.4}
\begin{tabular}{@{} l l @{}}
\toprule
\textbf{Dataset Feature} & \textbf{Details} \\
\midrule
Total Image-Text Pairs & $\sim$15.5 Million \\
Total Source Papers & $\sim$2.5 Million \\
Total Storage Size & 3.03 TB \\
Avg. Raw Caption Length (Mean $\pm$ SD) & $267 \pm 261$ characters \\
Avg. Enhanced Caption Length (Mean $\pm$ SD) & $759 \pm 251$ characters \\
Disciplinary Coverage & Physics, CS, Biology, Mathematics, Engineering, etc. \\
Data Sources & arXiv, bioRxiv, medRxiv, ChemRxiv, Nature Comms. \\
Data Format & JSONL (Metadata) \& TAR (Image Archives) \\
License & CC-BY-4.0 \\
\bottomrule
\end{tabular}
\caption{Overview of S1-MMAlign dataset specifications.}
\label{tab:key_info}
\end{table}

\begin{figure}[H]
    \centering
    \includegraphics[width=0.95\textwidth]{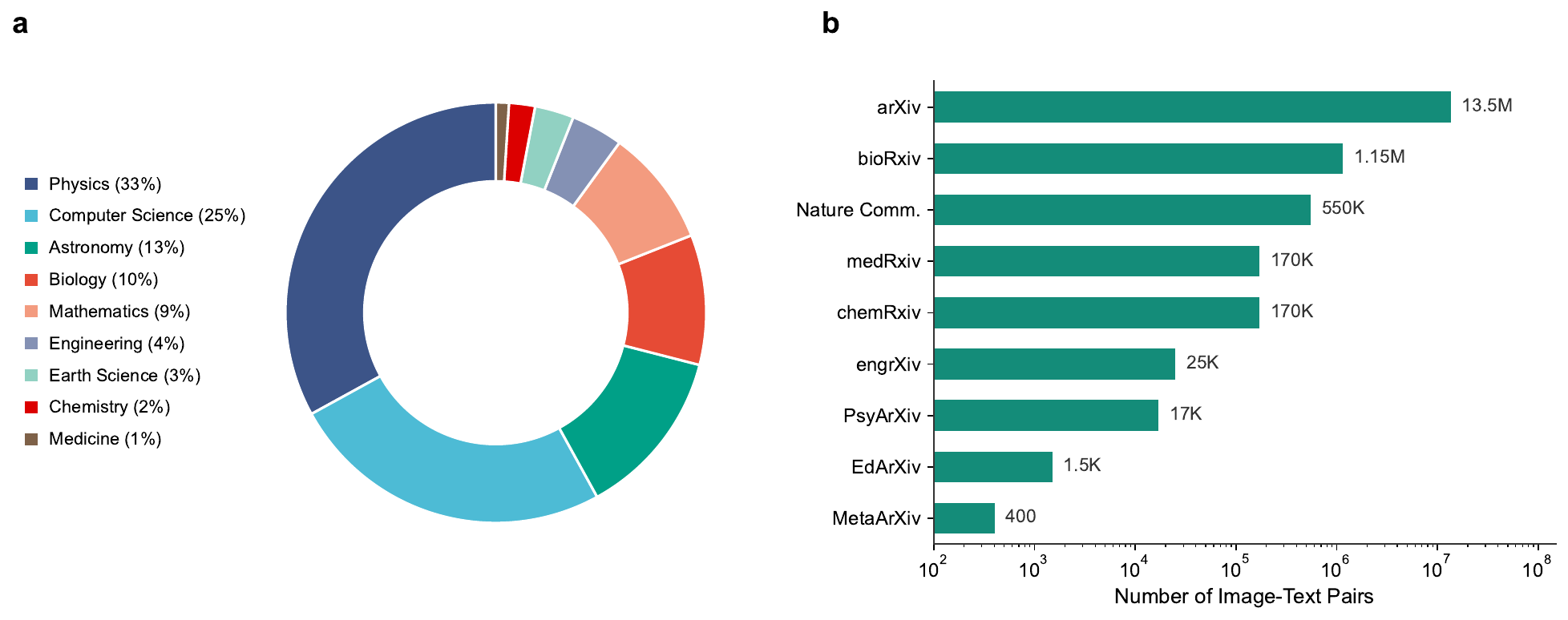}
    \caption{\textbf{Composition and scale of S1-MMAlign.} \textbf{a,} A ring chart presenting the proportional distribution of the dataset across diverse scientific domains. While quantitative disciplines (Physics, Computer Science) form the majority, the inclusion of natural and empirical sciences ensures a rich diversity of visual taxonomies, from charts to microscopic imagery. \textbf{b,} The distribution of extracted image-text pairs across source repositories, displayed on a logarithmic scale. The integration of massive-scale preprints (e.g., arXiv, bioRxiv) with high-quality peer-reviewed sources (e.g., Nature Communications) and niche archives enhances the model's generalization across heterogeneous scientific contexts.}
    \label{fig:subject_dist}
\end{figure}

To characterize the comprehensive composition of S1-MMAlign, we analyzed both the origins of the multimodal data and its subsequent disciplinary diversity. \textbf{Figure \ref{fig:subject_dist}} illustrates the distribution of the corpus across aggregated scientific domains and raw data sources, ensuring a holistic representation of scientific knowledge.

As illustrated by the proportional breakdown in \textbf{Figure \ref{fig:subject_dist}a}, the subject distribution is heavily anchored in quantitative disciplines. Physics and Computer Science dominate the ring chart, jointly comprising over half of the dataset and serving as the primary engines for multimodal scientific data. This robust quantitative core is complemented by significant contributions from natural sciences, such as Astronomy and Biology, and extends into a diverse long tail of empirical disciplines. This structural balance ensures comprehensive coverage across a wide spectrum of visual taxonomies.

Beyond subject aggregation, \textbf{Figure \ref{fig:subject_dist}b} details the provenance and immense scale of the extracted image-text pairs. The corpus is primarily driven by the massive foundations of arXiv and bioRxiv, yielding millions of pairs. Crucially, this preprint baseline is enriched with high-quality, peer-reviewed data from Nature Communications, alongside specialized repositories like medRxiv. The logarithmic scale further emphasizes our systematic effort to incorporate highly niche repositories—such as MetaArXiv—preventing domain overfitting and fostering a robust foundation that captures both mainstream and highly specialized scientific figures.

\section*{Methods}

The construction of S1-MMAlign follows a rigorous three-phase pipeline: multi-source data acquisition, robust image-text pair extraction, and semantic enhancement via context-aware recaptioning.
\subsection*{Data Sources}
To ensure a holistic representation of scientific knowledge, we implemented a systematic data acquisition protocol spanning diverse open-access repositories. Our harvesting strategy targets the primary preprint platforms—specifically arXiv, bioRxiv \cite{sever2019biorxiv}, medRxiv, and ChemRxiv—as well as the open-access subsets of high-impact journals such as Nature Communications. This multi-source approach ensures comprehensive disciplinary coverage, capturing a broad spectrum of visual modalities across the entire scientific landscape.

\subsection*{Image-Text Pair Extraction}

\subsubsection*{TeX Source Parsing}

We leveraged the rich semantic structure inherent in source code by retrieving raw TeX source packages via the arXiv bulk data access protocol. A custom extraction pipeline was implemented to handle the heterogeneity of user-uploaded sources. The process began with a rigorous integrity verification of the retrieved bundles; archives that were corrupted, failed decompression, or lacked essential image directories were systematically excluded, resulting in a rejection rate of 20\%. Following validation, we employed a robust parsing module based on Regular Expressions to navigate the TeX syntax. This module extracted \texttt{figure} environments to establish precise mappings between image filenames and their corresponding semantic metadata, including captions, labels, and local referencing contexts.

Visual assets underwent a standardization process to ensure compatibility with the input requirements of mainstream vision encoders \cite{radford2021learning}. Since scientific publications frequently utilize vector graphics (EPS/PDF), we implemented a rasterization step to convert these assets into high-quality PNG formats. Finally, a multi-stage heuristic filter was applied to eliminate noise and prioritize high-information scientific figures. This quality control stage discarded files smaller than 5KB, corrupted bytestreams, and non-semantic visual content, such as pure TeX tables rendered as images.

\subsubsection*{General Document Extraction}
Given that raw TeX source files are generally unavailable for repositories beyond arXiv (e.g., bioRxiv, medRxiv), we developed a custom extraction workflow anchored by the MinerU intelligent document processing engine \cite{mineru2024} to process the compiled PDF documents. This pipeline addresses the challenge of parsing unstructured layouts through a multi-stage approach described below.

\noindent\textbf{Deep Layout Detection.} We utilized MinerU's advanced layout analysis models to parse the document structure. This step precisely identifies and localizes key document elements—specifically figure boundaries and text blocks—generating precise bounding box coordinates for each component.

\noindent\textbf{Geometry-based Association.} Since layout detection outputs independent elements, we implemented a custom spatial proximity matching algorithm to link figures with their corresponding captions. By leveraging the coordinate information provided by MinerU, this heuristic logic associates image regions with the nearest valid caption text, effectively resolving figure-caption pairs even in complex multi-column layouts.

\noindent\textbf{Content Extraction and Cleaning.} Based on the detected coordinates, figure regions were extracted directly from the PDF source files. The associated caption text was simultaneously parsed and cleaned to remove non-semantic artifacts (e.g., indexing numbers). To ensure data quality, we implemented a post-extraction filtering protocol: extracted visual assets smaller than 5KB or images exhibiting rendering anomalies (such as solid black outputs caused by decoding failures) were automatically discarded, yielding structured image-text pairs aligned with the TeX-sourced data.

\subsection*{Semantic Enhancement via Context-Aware Recaptioning}
The core challenge in scientific multimodal learning lies in the "semantic gap" between raw visual signals and their high-level theoretical implications. A standard visual encoder might perceive a chart merely as geometric lines, whereas scientific reasoning requires interpreting these lines as manifestations of specific physical laws or experimental trends. To bridge this gap and generate descriptions that are not only visually accurate but also theoretically grounded, we implemented a semantic enhancement pipeline using the Qwen3-VL architecture \cite{qwen3vl2025}. We selected this model for its specialized SigLIP-2 encoder \cite{tschannen2025siglip2}, which utilizes 2D-RoPE to process dynamic high-resolution inputs. This capability is critical for preserving fine-grained scientific details—such as logarithmic axis scales, chemical bond structures, and error bars—that are often lost in standard low-resolution processing.

\noindent\textbf{Knowledge-Augmented Context Incorporation.} To ground visual perception in scientific narrative, we implemented a rigorous context-injection strategy. Rather than relying on generic captioning instructions, we constructed a holistic semantic window for each figure, integrating the paper's Title, Abstract, and the Local Citation Context—the specific text surrounding the figure's reference. This multi-source input compels the model to function as a scientific interpreter rather than a passive observer. By synthesizing the global research objective (from the abstract) with the "local" experimental analysis (from the citation context), the model generates captions that explain why the visual pattern exists, explicitly linking pixel-level features to domain-specific theoretical constructs. This approach effectively minimizes hallucinations and ensures the generated descriptions capture the underlying scientific causality.

\noindent\textbf{High-Throughput Inference Optimization.} Processing 15.5 million scientific figures required a robust computational framework. We deployed a parallelized inference pipeline on an 8 $\times$ H100 GPU cluster, leveraging the vLLM library \cite{kwon2023efficientmemorymanagementlarge} to handle the massive throughput requirements. By utilizing PagedAttention for efficient memory management and continuous batching strategies, we maximized hardware utilization and token generation speed. This optimized infrastructure enabled the efficient transformation of the entire corpus at scale, converting sparse raw images into a dense, knowledge-rich multimodal dataset ready for downstream scientific reasoning tasks.

\section*{Data Records}
The S1-MMAlign dataset is hosted on the Hugging Face Hub. To facilitate modular access and efficient subsetting across diverse scientific domains, the corpus is organized into a stratified hierarchy based on source provenance (e.g., arXiv, bioRxiv).

\subsection*{File Organization}
The repository implements a decoupled storage architecture designed to optimize the retrieval and processing of high-volume multimodal data. As illustrated in Figure \ref{fig:file_structure}, this architecture separates lightweight semantic metadata from heavy visual assets to support flexible access patterns.

\noindent\textbf{Metadata Serialization.} Semantic annotations, bibliographic metadata, and alignment information are serialized in the \texttt{JSONL} (JSON Lines) format and housed within the \texttt{jsonl/} directory. This line-oriented standard was specifically selected to support large-scale streaming and compatibility with distributed processing frameworks (e.g., Apache Spark), allowing researchers to parse text-based features without the latency of loading binary image data.

\noindent\textbf{Sharded Visual Archival.} Corresponding visual content (e.g., plots, diagrams) is maintained in standard formats (PNG/JPEG) and packaged into compressed \texttt{.tar.gz} archives. To mitigate I/O bottlenecks and ensure download stability for terabyte-scale subsets, we adopted a multi-volume sharding strategy: archives exceeding 30GB are segmented into sequential parts (e.g., \texttt{images.tar.gz.partaa}). This segmentation facilitates parallelized data transfer and allows for robust recovery during interrupted downloads.

\noindent\textbf{Cryptographic Integrity Verification.} To guarantee data reliability and optimize storage efficiency for redundant visual assets, the repository is managed via the Xet version control extension \cite{low2023git}. Unlike traditional Git-LFS, Xet utilizes Merkle-tree based deduplication, which segments large files into content-addressable blocks. The system exposes a dual-layer verification mechanism: a SHA-256 checksum ensures the bit-level consistency of the reconstructed file content, while the Xet Hash validates the structural integrity of the stored blocks. Researchers can utilize these cryptographic fingerprints (contained within the 135-byte pointer files) to algorithmically verify that downloaded artifacts are exact, corruption-free replicas of the source.

\begin{figure}[H]
    \centering
    \renewcommand*\DTstylecomment{\rmfamily\color{gray}\textsc} 
    \renewcommand*\DTstyle{\ttfamily\textcolor{black}\small} 
    \setlength{\DTbaselineskip}{15pt} 

    \begin{minipage}{0.95\textwidth}
    \dirtree{%
    .1 S1-MMAlign/ \DTcomment{Root Repository}.
    .2 arxiv/ \DTcomment{Structure Type A: Large-scale split archives}.
    .3 jsonl/ \DTcomment{Contains metadata files}.
    .4 07\_recaption.jsonl.
    .4 08\_recaption.jsonl.
    .4 ....
    .4 25\_recaption.jsonl.
    .3 images\_2007.tar.gz 
    .3 images\_2008.tar.gz.
    .3 images\_2009.tar.gz.partab.
    .3 ....
    .2 biorxiv/ \DTcomment{Structure Type B: Multi-part archives}.
    .3 jsonl/.
    .4 biorxiv\_recaption.jsonl.
    .3 images.tar.gz.partaa.
    .3 images.tar.gz.partab.
    .3 ....
    .2 chemrxiv/ \DTcomment{Structure Type C: Single archive}.
    .3 jsonl/.
    .4 chemrxiv\_recaption.jsonl.
    .3 images.tar.gz.
    .2 {[nature\_communications]} \DTcomment{Type B}.
    .2 {[edrxiv, engrxiv, medrxiv, metarxiv, psyarxiv]} \DTcomment{Type C}.
    .2 README.md.
    .2 .gitattributes.
    }
    \end{minipage}
    \caption{File organization of S1-MMAlign. The repository structure adapts to data volume: (A) Yearly Archives for massive sources like arXiv (e.g., \texttt{images\_2007.tar.gz}); (B) Multi-Part Archives for large sources like bioRxiv (e.g., \texttt{images.tar.gz.partaa}); and (C) Single Archives for smaller datasets. All subsets include a \texttt{jsonl} directory for metadata.}
    \label{fig:file_structure}
\end{figure}

\subsection*{Metadata Schema}
The dataset follows a flat JSON structure where each key corresponds to a specific semantic attribute. The detailed schema definition and data types are presented in Table \ref{tab:schema}.

\begin{table}[htbp]
\centering
\renewcommand{\arraystretch}{1.3}
\begin{tabular}{@{} l p{10.5cm} @{}}
\toprule
\textbf{Field Key} & \textbf{Description} \\
\midrule
\texttt{doi} / \texttt{arxiv\_id} & Unique identifier for the source publication (DOI or arXiv ID). \\
\texttt{title} & Title of the source scientific paper. \\
\texttt{image\_path} & Relative file path pointing to the image file. \\
\texttt{caption} & Raw figure caption extracted directly from the source. \\
\texttt{recaption} & Semantically enhanced description generated by the pipeline. \\
\texttt{categories} & Disciplinary classification. \\
\bottomrule
\end{tabular}
\caption{Metadata schema definitions. Description of semantic attributes present in each JSONL record.}
\label{tab:schema}
\end{table}
\section*{Technical Validation}
To rigorously quantify the efficacy and broad utility of S1-MMAlign, we implemented a multi-dimensional validation protocol spanning \textbf{intrinsic data analysis} and \textbf{extrinsic model evaluation}. Our assessment proceeds in three systematic stages: 
(1) \textbf{Data-Centric Quality Verification}, focusing on information density, linguistic fluency (SciBERT), and cross-modal alignment (CLIP); 
(2) \textbf{Unlocking Scientific Capabilities}, verifying the dataset's ability to drive zero-shot generative generalization and complex scientific reasoning via the Qwen3-VL framework; and 
(3) \textbf{Empowering MLLM Pre-training}, demonstrating the dataset's effectiveness in boosting domain-specific continual pre-training and visual instruction tuning across diverse model architectures (LLaVA-1.5/NeXT) on 10 scientific QA tasks.

\subsection*{Data-Centric Quality Verification}
\subsubsection*{Textual Expansion and Context Enrichment}
We first quantified the extent of textual expansion achieved by our enhancement pipeline by analyzing the character length distribution across the corpus. Scientific figures often suffer from semantic sparsity, where raw captions provide insufficient context for training multimodal models. 

\begin{figure}[H]
    \centering
    \includegraphics[width=0.85\textwidth]{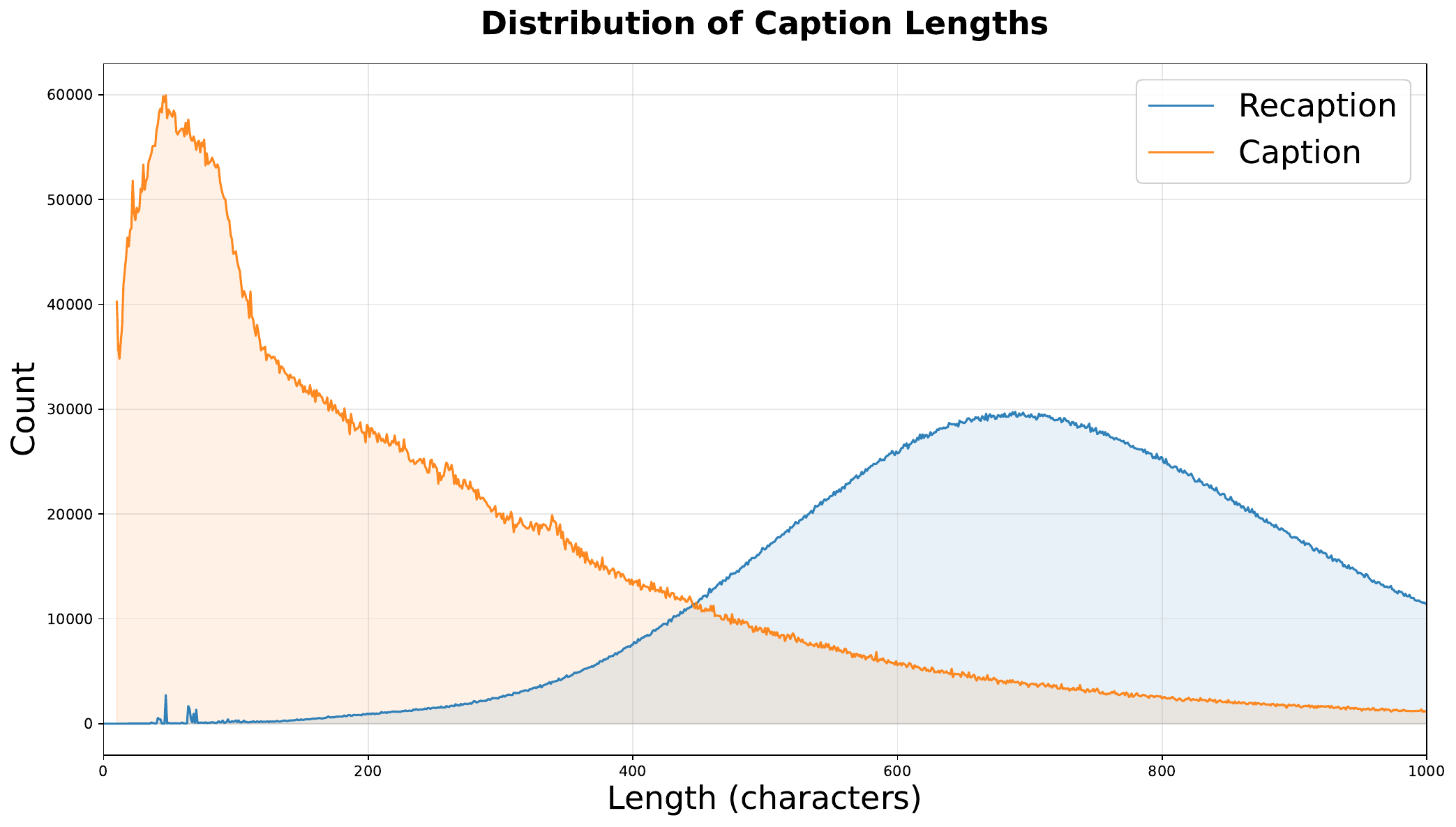}
    \caption{Character length distribution analysis. The comparison between raw (orange) and enhanced (blue) corpora highlights the substantial contextual enrichment achieved by our pipeline. The enhancement process not only expands the textual content by approximately 2.8 times but also reduces length variance, effectively mitigating the semantic sparsity inherent in original scientific metadata.}
    \label{fig:length_dist}
\end{figure}

As demonstrated in Figure \ref{fig:length_dist}, our recaptioning strategy effectively bridges this gap. Comparative statistics reveal a significant shift towards more comprehensive descriptions: raw captions exhibit high volatility with a character count of $267 \pm 261$ (mean $\pm$ std). In contrast, the semantically enhanced descriptions  achieve a robust 2.8$\times$ expansion, yielding $759 \pm 251$ characters. The significant reduction in the coefficient of variation (from $97.8\%$ to $33.1\%$) indicates a more homogeneous and standardized semantic representation, providing a consistently rich contextual foundation for the subsequent quality assessments.

\subsubsection*{Linguistic Fluency via Pseudo-Perplexity}
We employed SciBERT \cite{beltagy2019scibert} to compute the pseudo-Perplexity (pseudo-PPL) of the generated captions, serving as a robust proxy for linguistic fluency and domain adaptation. Unlike general-purpose language models, SciBERT is pre-trained specifically on scientific corpora, making it uniquely sensitive to the syntactic structures and terminologies inherent in academic discourse. Lower pseudo-PPL values indicate higher sequence probability, reflecting text that is more linguistically coherent and aligned with standard scientific language patterns \cite{devlin2018bert}.

\begin{figure}[H]
    \centering
    \includegraphics[width=0.8\textwidth]{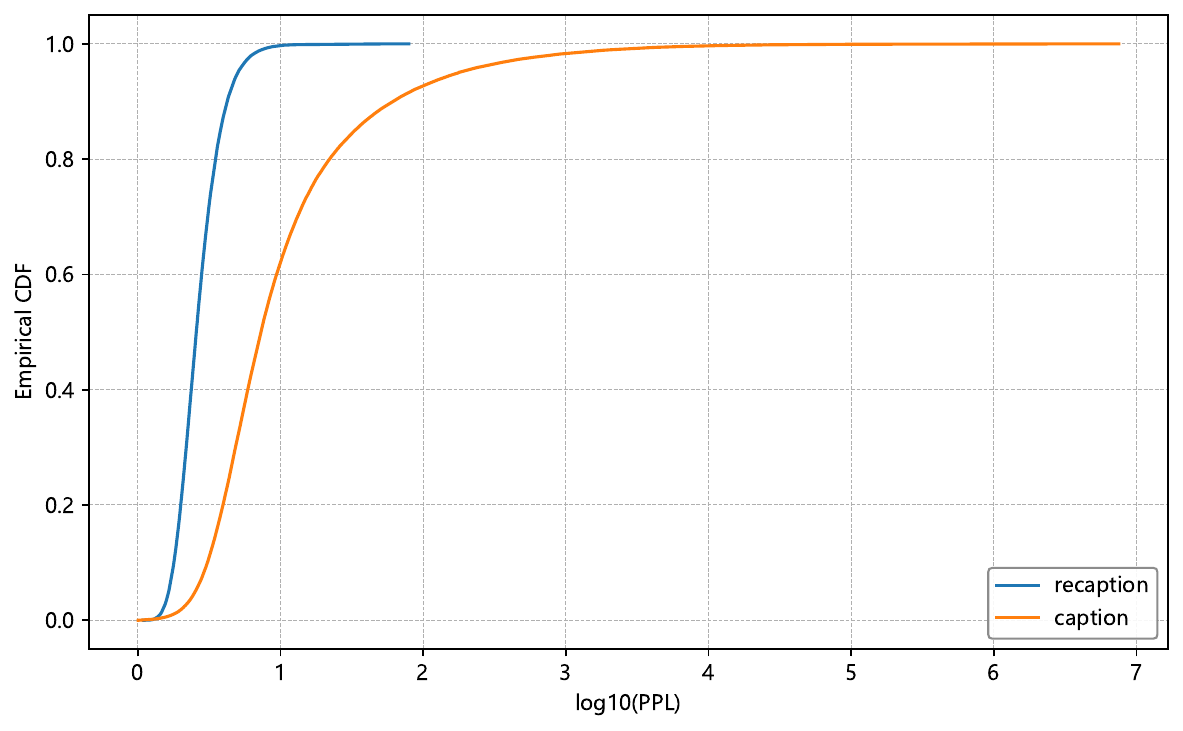}
    \caption{Empirical CDF of text quality (Pseudo-PPL). The plot illustrates the Cumulative Distribution Function of $\log_{10}(\text{pseudo-PPL})$ scores derived from SciBERT \cite{beltagy2019scibert}. The blue curve (Enhanced Captions) demonstrates a pronounced leftward shift compared to the original captions, confirming a significant reduction in perplexity and superior alignment with scientific linguistic norms.}
    \label{fig:ppl}
\end{figure}

As visualized in Figure \ref{fig:ppl}, the distribution of enhanced captions exhibits a significant \textbf{leftward shift} towards lower perplexity values. This statistical trend confirms that the enhancement process effectively mitigates ambiguity and improves the syntactic completeness of the descriptions, resulting in higher-fidelity scientific text.

\subsubsection*{Image-Text Semantic Alignment and Consistency}
The CLIP Score metric, rooted in cosine similarity embeddings \cite{radford2021learning, li2022blip}, was employed to rigorously quantify cross-modal alignment between visual content and textual descriptions. Specifically, we conducted a comparative analysis between the original raw captions and our semantically enhanced counterparts to validate the efficacy of the recaptioning pipeline.

\noindent\textbf{Quantitative Alignment Improvement.} The evaluation reveals a substantial quality uplift, with the mean CLIP score increasing by \textbf{18.21\%}. This metric shift indicates that the context-aware recaptioning generates descriptions with significantly higher semantic congruity to the visual data, successfully recovering visual details often omitted in sparse original captions.

\noindent\textbf{Distributional Homogeneity.} Beyond mean improvements, the variance of the score distribution decreased by approximately \textbf{27.77\%}. This reduction in volatility signifies improved quality consistency across the corpus. It suggests that the pipeline effectively mitigates low-quality outliers, transforming a highly variable raw dataset into a standardized corpus with stable cross-modal alignment.

\begin{figure}[H]
    \centering
    \includegraphics[width=0.8\textwidth]{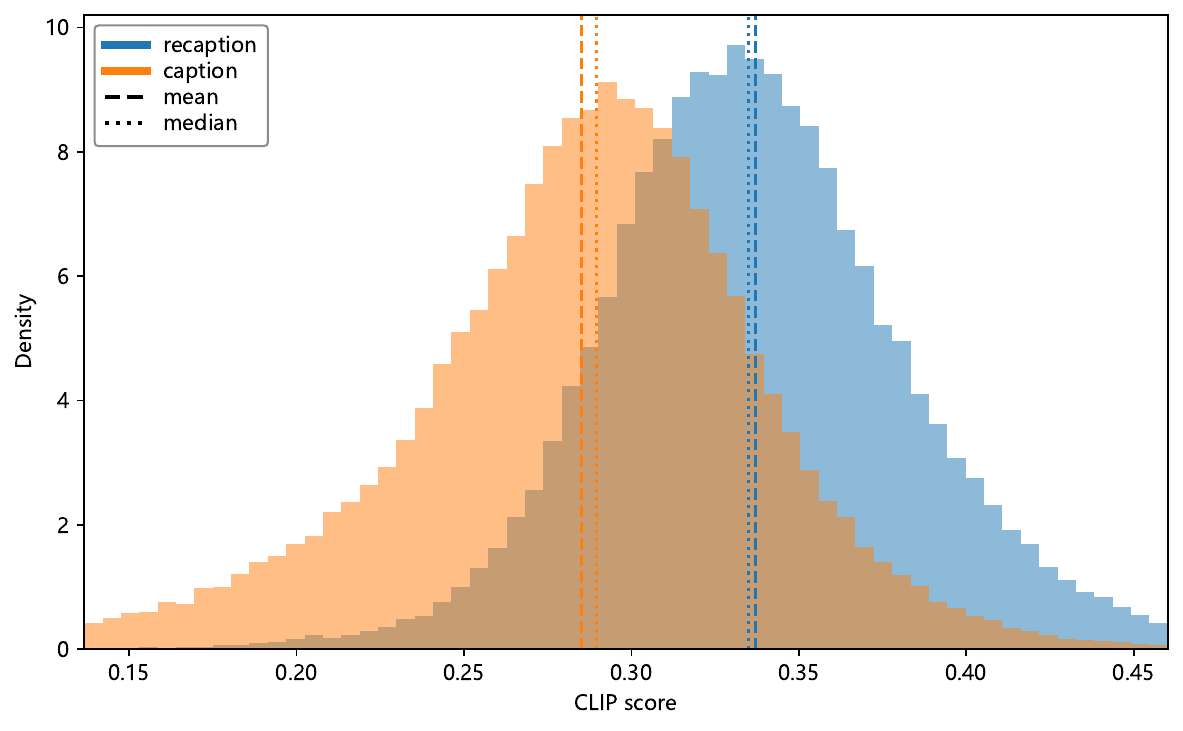}
    \caption{Distribution of CLIP image-text consistency scores. The histogram compares the alignment scores of original (orange) versus enhanced (blue) captions. The pronounced rightward shift and narrower spread of the blue distribution indicate that the enhancement strategy yields a dataset with higher semantic fidelity and greater consistency, providing a stronger supervision signal for multimodal representation learning.}
    \label{fig:clip}
\end{figure}

As visualized in Figure \ref{fig:clip}, the distribution for enhanced captions is clearly distinguished from the baseline. The observed rightward shift confirms that the recaptioned text provides a much stronger and more reliable supervision signal for downstream multimodal training tasks.

\subsection*{Unlocking Scientific Generation and Reasoning Capabilities}

To rigorously assess the quality of the textual descriptions provided by S1-MMAlign, we fine-tuned a \textit{Qwen3-VL-2B} model \cite{qwen3vl2025} on a subset of our dataset (10M pairs). We designed a dual-aspect evaluation protocol to verify both the linguistic quality of image captioning and the model's effectiveness in Visual Question Answering (VQA).

\subsubsection*{Zero-Shot Generalization in Scientific Captioning}

First, we evaluated the model's generative capabilities on the SciCap test set \cite{hsu2021scicap}. Crucially, this evaluation was conducted in a zero-shot setting: the model was trained solely on S1-MMAlign without exposure to SciCap's training data. This setup rigorously tests the dataset's generalization ability and whether the learned semantic alignment is transferable to unseen scientific figures.

\begin{figure}[ht]
    \centering
    \includegraphics[width=0.8\textwidth]{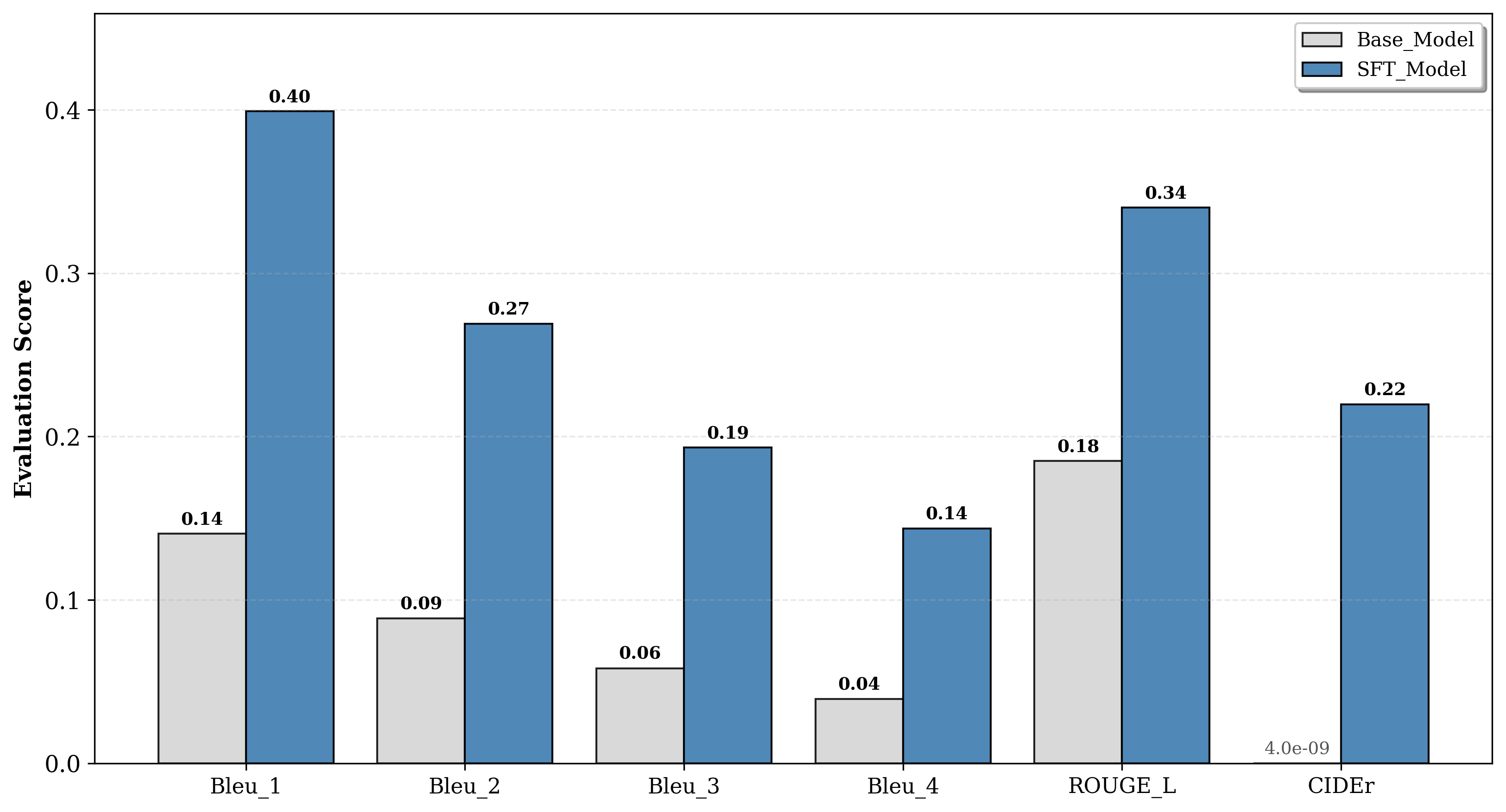}
    \caption{Zero-Shot generative performance on SciCap benchmark. Comparison of standard captioning metrics. The dramatic increase in BLEU and CIDEr scores in a zero-shot setting confirms the dataset's efficacy in aligning visual features with precise scientific terminology.}
    \label{fig:scicap_metrics}
\end{figure}

As shown in Figure \ref{fig:scicap_metrics}, the model fine-tuned on S1-MMAlign achieves a significant performance leap over the baseline. Notably, the CIDEr score \cite{vedantam2015cider}, a metric critical for measuring consensus with human-generated captions, rose from negligible to 0.22, while BLEU-4 scores \cite{papineni2002bleu} improved from 0.04 to 0.14. These metrics confirm that S1-MMAlign effectively teaches the model to synthesize scientifically accurate and structurally coherent descriptions, demonstrating strong generalization potential beyond the source training distribution.

\subsubsection*{Scientific Reasoning via Visual Question Answering}

Beyond linguistic fluency, we assessed whether the enhanced captions contain sufficient information to support complex scientific reasoning. We employed a frozen-evaluator paradigm: the fine-tuned \textit{Qwen3-VL-2B} generated captions for images in various benchmarks, which were then fed into a frozen \textit{Qwen3-VL-30B} model to answer questions \textit{without} seeing the original image. This setup isolates the information density of the captions as the sole variable. As presented in Table \ref{tab:downstream_results}, this evaluation reveals a distinct capability shift induced by S1-MMAlign across a wide spectrum of scientific reasoning tasks.

\begin{table}[ht]
\centering
\resizebox{0.65\textwidth}{!}{%
\begin{tabular}{lcc}
\toprule
\textbf{Benchmark} & \multicolumn{2}{c}{\textbf{Qwen3-VL-2B}} \\
\cmidrule(lr){2-3}
& \textbf{Base} & \textbf{+ S1-MMAlign (Ours)} \\
\midrule
MMMU & 46.89\% & \textbf{49.67\%} \\
ChartQA & 52.80\% & \textbf{53.72\%} \\
ScienceQA & 86.51\% & \textbf{86.76\%} \\
PhysUniBench & 9.16\% & \textbf{10.18\%} \\
MathVista & 45.70\% & \textbf{46.10\%} \\
ArxivQA & 63.40\% & \textbf{67.90\%} \\
EMMA & 19.08\% & \textbf{23.31\%} \\
PMC-VQA & 43.60\% & \textbf{47.65\%} \\
SciFIBench & 61.38\% & \textbf{67.80\%} \\
\bottomrule
\end{tabular}%
}
\caption{Downstream performance verification using generated captions. We compare the generative quality of the Base model versus the model fine-tuned on S1-MMAlign. Evaluation uses a frozen evaluator to assess caption utility across diverse benchmarks. The best performance for each benchmark is highlighted in bold.}
\label{tab:downstream_results}
\end{table}

\textbf{Gains in Scientific Reasoning.} The model achieves substantial performance improvements in knowledge-intensive benchmarks, including SciFIBench \cite{roberts2024scifibench} (\textbf{+6.42\%}), ArxivQA \cite{li2024multimodal} (\textbf{+4.50\%}), EMMA \cite{hao2025emma} (\textbf{+4.23\%}), PMC-VQA\cite{zhang2023pmc} (\textbf{+4.05\%}), and MMMU \cite{yue2023mmmu} (\textbf{+2.78\%}). These tasks require interpreting the \textit{implications} of scientific figures (e.g., experimental conclusions, model architectures) rather than merely perceiving visual patterns. The results demonstrate that S1-MMAlign effectively bridges the semantic gap, enabling the model to extract and describe critical scientific logic that is essential for high-level reasoning.

In summary, the significant gains across these multidisciplinary reasoning benchmarks confirm that S1-MMAlign successfully equips models with the domain-specific knowledge required for "AI for Science" applications.

\subsection*{Empowering Scientific MLLMs via Enhanced Pre-training}

To demonstrate the broad applicability of S1-MMAlign in end-to-end visual instruction tuning, we conducted extensive evaluations using the \textit{LLaVA-1.5} \cite{liu2023improvedllava} and \textit{LLaVA-NeXT} \cite{li2024llavanext-strong} architectures. We implemented a rigorous \textbf{three-stage curriculum learning strategy} to systematically assess the impact of data quality on model performance. Unlike standard two-stage protocols, we introduced an intermediate scientific adaptation stage:

\begin{description}
    \item[Stage 1 (Modality Alignment):] Training the projector (MLP) on the official 558k pre-training dataset to establish initial visual-text alignment.
    
    \item[Stage 2 (Scientific Adaptation):] Training the projector and LLM on 1 million image-text pairs from our S1-MMAlign dataset. Here, we specifically compared models trained with raw "Caption" versus our enhanced "Recaption" to validate the efficacy of domain-specific knowledge injection.
    
    \item[Stage 3 (Supervised Fine-Tuning):] Final instruction tuning on the official LLaVA SFT dataset to equip the model with conversational capabilities.
\end{description}

\textbf{Architectural Configurations:} To ensure a controlled comparison, both frameworks utilized identical backbones: a Qwen2.5-3B language model \cite{qwen25} paired with a SigLIP-SO400M vision encoder \cite{tschannen2025siglip2}. The key distinction lies in the training strategy. For \textit{LLaVA-1.5}, the vision encoder remained frozen throughout all three training stages, with updates restricted to the projector and LLM components. In contrast, for \textit{LLaVA-NeXT}, we leveraged its dynamic high-resolution (AnyRes) architecture. Crucially, during Stage 3, we performed full-parameter fine-tuning (unfreezing the Vision Encoder, Projector, and LLM). This strategy allows the visual encoder to adaptively process fine-grained details in scientific charts and dense document layouts.

\textbf{Results Analysis.} Table \ref{tab:llava_results} summarizes the performance across 10 diverse scientific benchmarks. The integration of S1-MMAlign (Recaption) yields consistent improvements across the majority of evaluations, validating the dataset's broad utility.

\textbf{Visual Parsing Dominance.} The most pronounced gains are observed in tasks requiring dense visual extraction and layout understanding. For \textit{LLaVA-NeXT}, performance on ChartQA \cite{masry2022chartqa} surged from 30.24\% to \textbf{51.88\%}, and DocVQA \cite{mathew2021docvqa} improved from 42.89\% to \textbf{58.00\%}. Similar trends are evident in the LLaVA-1.5 group, confirming that the full-parameter tuning coupled with our data effectively unlocks the model's ability to resolve fine-grained chart details and complex document structures that are often missed by standard training.

\textbf{Enhanced Scientific Reasoning.} Beyond visual perception, the model demonstrates stronger domain-specific reasoning capabilities. Significant uplifts were recorded in SciFIBench (\textbf{+11.6\%} for NeXT), ArxivQA (\textbf{+7.3\%}), and MMMU (\textbf{+3.9\%}). This indicates that the recaptions not only describe \textit{what} is in the image but also encapsulate the \textit{scientific context}—such as experimental conditions and data trends—necessary for solving complex, multidisciplinary problems.

\textbf{Broad Domain Generalization.} The comprehensive improvement observed across specialized disciplines (e.g., PhysUniBench \cite{wang2025physunibench}, PMC-VQA) confirms that the model has acquired richer domain knowledge. By consistently outperforming the raw Caption baseline across both architectures, the Recaption strategy proves to be a superior data source for scientific multimodal learning.

\begin{table}[htbp]
\centering
\resizebox{\textwidth}{!}{%
\begin{tabular}{l|ccc|ccc}
\toprule

\multirow{2}{*}{} & \multicolumn{3}{c|}{LLaVA-1.5 } & \multicolumn{3}{c}{LLaVA-NeXT } \\
\cmidrule(lr){2-4} \cmidrule(lr){5-7}
 & Base & +Caption & +S1-MMAlign & Base & +Caption & +S1-MMAlign \\
\midrule
MMMU & \textbf{36.56} & 33.00 & 36.22 & 35.44 & 37.33 & \textbf{39.33} \\
ChartQA & 18.00 & 20.40 & \textbf{33.08} & 30.24 & 39.72 & \textbf{51.88} \\
PhysUniBench & 27.99 & 30.79 & \textbf{34.86} & 19.08 & 21.37 & \textbf{25.45} \\
MathVista & 30.60 & 30.40 & \textbf{31.10} & 33.10 & 33.40 & \textbf{34.30} \\
DocVQA & 23.78 & 27.48 & \textbf{34.37} & 42.89 & 51.43 & \textbf{58.00} \\
ArxivQA & 55.00 & 57.70 & \textbf{61.20} & 55.40 & 59.20 & \textbf{62.70} \\
EMMA & 18.69 & 19.26 & \textbf{19.69} & 15.85 & 18.26 & \textbf{18.83} \\
PlotQA & 5.05 & 6.35 & \textbf{7.75} & 7.45 & 8.45 & \textbf{10.50} \\
PMC-VQA & 37.80 & 39.95 & \textbf{42.05} & 38.90 & 40.05 & \textbf{40.75} \\
SciFIBench & 30.40 & 35.40 & \textbf{36.40} & 34.20 & 45.00 & \textbf{45.80} \\
\bottomrule
\end{tabular}%
}
\caption{Impact of S1-MMAlign on downstream scientific benchmarks. By transposing the layout, we provide a clearer comparison across 10 datasets. The best performance within each model architecture group (LLaVA-1.5 and LLaVA-NeXT) is highlighted in bold. Our Recaption strategy yields consistent improvements, especially on visually complex tasks.}
\label{tab:llava_results}
\end{table}

\section*{Usage Notes}
The dataset is designed for pre-training scientific multimodal models \cite{10483603}. We recommend using the Hugging Face \texttt{datasets} library to stream the data, as downloading the entire 15.5 million set may require significant storage. Standard image processing libraries (e.g., PIL) can be used to load images from the provided paths.

Users should be aware of certain technical boundaries when utilizing this dataset. First, since the recaptioning pipeline relies on a VLM (Qwen-VL), there remains a risk of model hallucination, particularly in highly specialized sub-fields. Users training critical scientific models should consider further domain-specific validation. Second, the disciplinary distribution naturally reflects the open-access literature landscape, meaning physics and computer science are heavily represented. For fields like chemistry, researchers may need to incorporate supplementary data to achieve balanced downstream training.

\section*{Data Availability}

The S1-MMAlign dataset generated and analysed during the current study is publicly available in the Hugging Face repository at \url{https://huggingface.co/datasets/ScienceOne-AI/S1-MMAlign} and is permanently archived with the DOI: \url{https://doi.org/10.57967/hf/8008} \cite{he_wang_2026}.The dataset is distributed under the Creative Commons Attribution-NonCommercial 4.0 International (CC-BY-NC-4.0) license for research and non-commercial use only.

This repository contains the complete dataset comprising over 15.5 million image-text pairs. To optimize retrieval and storage, the dataset implements a decoupled storage architecture. Semantic metadata—including original scientific figures' DOIs, raw captions, and AI-enhanced captions—are serialized in JSON Lines (\texttt{.jsonl}) format. The corresponding visual assets (stored in standard PNG/JPEG formats) are packaged into compressed \texttt{.tar.gz} archives. To ensure download stability for massive data subsets, large visual archives are segmented into multi-volume shards (e.g., \texttt{.partaa}, \texttt{.partab}). Furthermore, data integrity can be cryptographically verified using the repository's Xet-based SHA-256 checksums.

\section*{Code Availability}

The custom codebase utilized for the construction and semantic enhancement of the S1-MMAlign dataset is publicly available in our Hugging Face repository at \url{https://huggingface.co/datasets/ScienceOne-AI/S1-MMAlign} and archived under the dataset's DOI \cite{he_wang_2026}. To facilitate transparency and reproducibility, the core data generation pipeline is organized into three main modules. Specifically, the \texttt{data\_download} module contains scripts for fetching open-access literature and managing initial PDF acquisition. The \texttt{data\_processing} module includes pipelines for document parsing (e.g., utilizing MinerU), figure-text extraction, and multi-stage data filtering. Furthermore, the \texttt{recaption} module provides the knowledge-augmented prompting templates and execution logic utilized with the Qwen-VL model series for semantic caption enhancement.

For the downstream model training and benchmark evaluations presented in the Technical Validation section, we employed standard, publicly available vision-language model training frameworks and community-driven evaluation suites. As these rely on established open-source infrastructures and do not represent custom software developed specifically for this data descriptor, their standard scripts are not duplicated in our repository.


\section*{Author Contributions}
H.W. and L.G. conceived the project, designed the semantic enhancement pipeline, led the data collection and processing, and conducted the technical validation. P.H., X.L., Y.Y., and J.J. contributed to the data processing. J.L. supervised the research and acquired funding. All authors contributed to the writing, reviewing, and editing of the manuscript.

\section*{Competing Interests}
The authors declare no competing interests.

\section*{Funding}
This work was supported by the Strategic Priority Research Program of Chinese Academy of Sciences [grant number XDB1350103]; the National Natural Science Foundation of China [grant numbers 62437001, 62436001]; and the Key Research and Development Program of Jiangsu Province [grant number BE2023016-3].

\clearpage
\begin{center}
    \LARGE \textbf{Supplementary Information}
\end{center}
\vspace{1em}

\noindent \textbf{Supplementary Table 1:} Detailed comparison of images, original captions, and recaptions.
\vspace{1em}

\begin{longtable}{
  >{\centering\arraybackslash}m{0.35\textwidth} 
  >{\raggedright\arraybackslash}m{0.6\textwidth}
}
\toprule
\textbf{Images} & \textbf{Captions} \\
\midrule
\endfirsthead

\toprule
\textbf{Images} & \textbf{Captions} \\
\midrule
\endhead

\includegraphics[width=\linewidth]{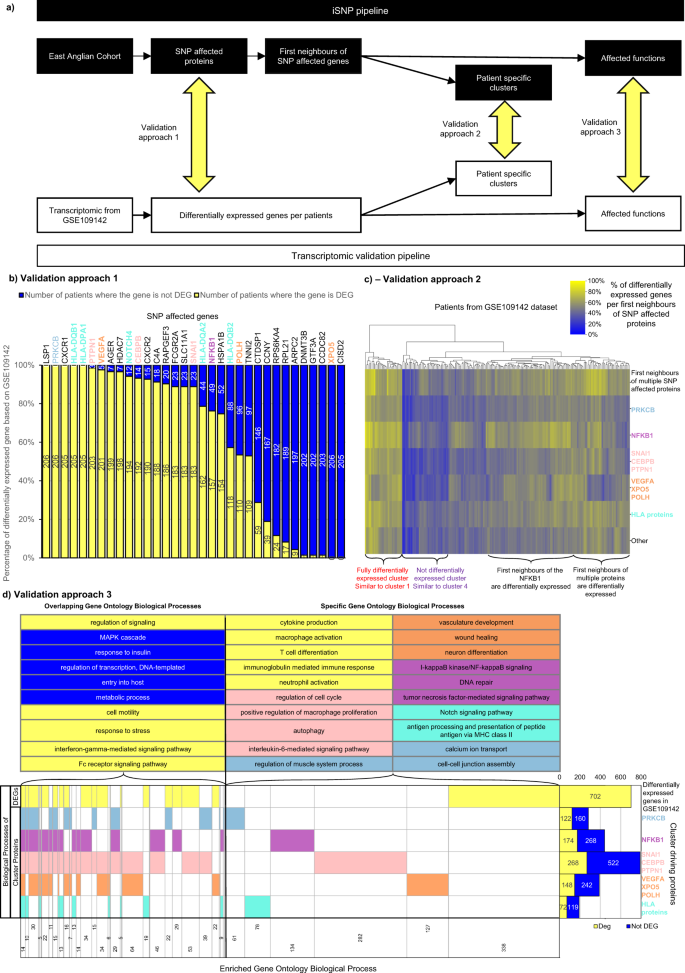} & 
\textbf{Original caption:} \newline 
Fig. 4: Validation of the iSNP method with transcriptomic data from an independent cohort of ulcerative colitis patients. 
\vspace{1em} 
\newline
\textbf{Recaption:} \newline
a composite scientific figure illustrating a validation pipeline for a systems genomics approach to ulcerative colitis. Panel (a) shows a flowchart of the iSNP pipeline, starting with SNP-affected proteins from the East Anglian Cohort, identifying their first neighbors, and generating patient-specific clusters and affected functions, which are then validated using transcriptomic data from GSE109142. Panel (b) is a bar chart showing the percentage of patients in which each SNP-affected gene is differentially expressed (DEG), with the number of patients where the gene is DEG indicated above each bar. Panel (c) is a heatmap showing the percentage of differentially expressed genes among the first neighbors of SNP-affected proteins in the GSE109142 dataset, with patients clustered into four groups based on their expression patterns. Panel (d) is a table and a stacked bar chart showing overlapping Gene Ontology Biological Processes between differentially expressed genes and first neighbors of cluster-driving proteins, with the number of overlapping processes indicated. \\
\midrule

\includegraphics[width=\linewidth]{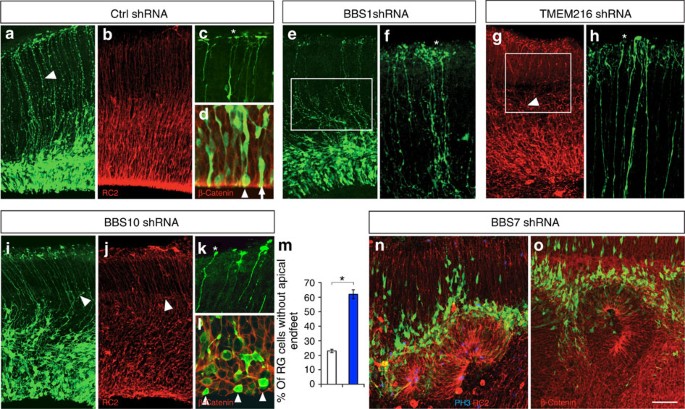} & 
\textbf{Original caption:} \newline 
Figure 2: Ciliopathy-related gene effects on polarized radial glial progenitor morphology.
\vspace{1em} 
\newline
\textbf{Recaption:} \newline
a series of fluorescent microscopy images and a bar chart comparing the morphology of radial glial (RG) progenitors in mouse embryonic cerebral cortex under different genetic conditions. The top row shows control conditions (Ctrl shRNA) and knockdowns of BBS1, TMEM216, BBS10, and BBS7 using shRNA. Each panel uses fluorescent markers to visualize RG cells: green for GFP-labeled RG processes, red for RC2 (a marker for radial glia), and red for \textbeta-catenin (a marker for the apical surface). The images show the organization of RG cells, including their basal processes (spanning the cortical width), basal endfeet (at the pial surface), and apical endfeet (at the ventricular surface). The control panels (a-d) show a well-organized scaffold with polarized RG cells. In contrast, the knockdown panels show various morphological disruptions: wavy basal processes and disrupted basal endfeet in BBS1 shRNA (e,f); reduced fasciculation and misorientation in TMEM216 shRNA (g,h); aberrantly branched, short, or retracted basal processes and loss of apical endfeet in BBS10 shRNA (i-l); and rosette-like organization with aberrant β-catenin expression in BBS7 shRNA (n,o). Panel m is a bar chart quantifying the percentage of RG cells without apical endfeet, showing a significantly higher percentage in BBS10 shRNA compared to control. Scale bars are provided for each panel. \\
\midrule

\includegraphics[width=\linewidth]{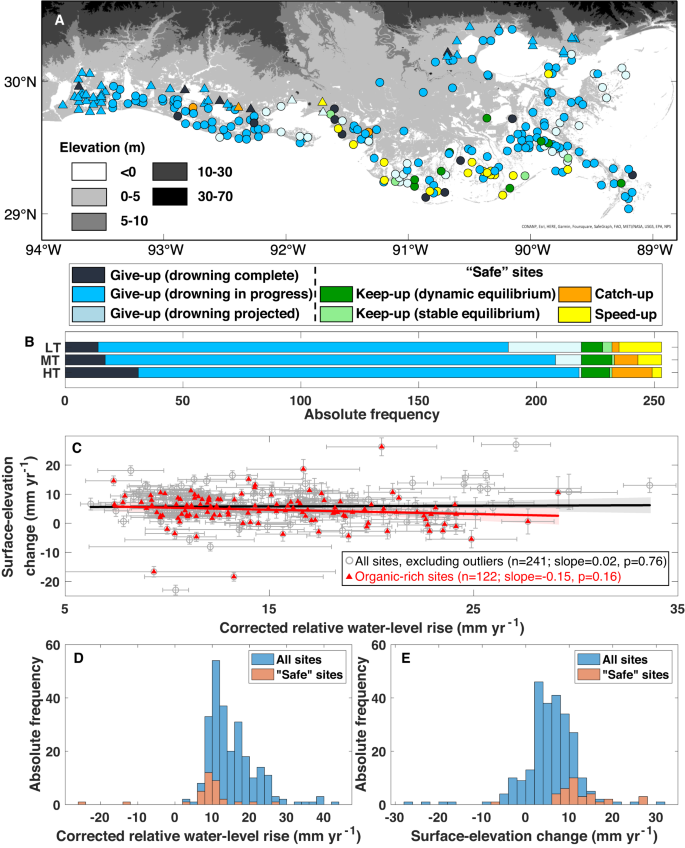} & 
\textbf{Original caption:} \newline 
Fig. 4: Wetland response to corrected relative water-level (cRWL) change at each monitoring site.
\vspace{1em} 
\newline
\textbf{Recaption:} \newline
a composite scientific figure with five panels (A--E) showing data on coastal wetland response to relative water-level rise. Panel A is a map of Louisiana's coastal region with monitoring sites plotted by color and symbol, indicating elevation and wetland response categories (e.g., ``Give-up,'' ``Keep-up,'' ``Speed-up'') based on surface-elevation change. Panel B is a stacked bar chart showing the absolute frequency of sites categorized by wetland response under low tide (LT), mean tide (MT), and high tide (HT) conditions. Panel C is a scatter plot of surface-elevation change (mm yr$^{-1}$) versus corrected relative water-level rise (mm yr$^{-1}$), with data points for all sites and a subset of organic-rich sites, along with linear regression lines and statistical information. Panel D is a histogram showing the distribution of corrected relative water-level rise rates for all sites and a subset labeled ``Safe'' sites. Panel E is a histogram showing the distribution of surface-elevation change rates for all sites and the ``Safe'' sites. \\
\midrule

\includegraphics[width=\linewidth]{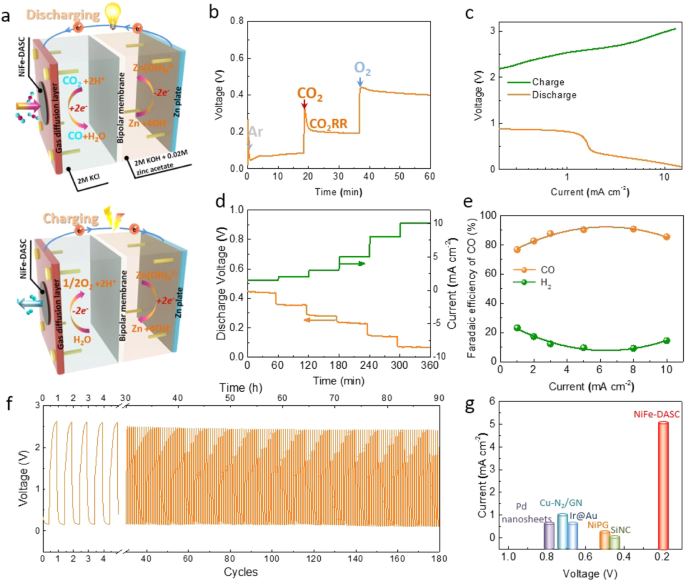} & 
\textbf{Original caption:} \newline 
Fig. 4: The rechargeable Zn-CO2battery cell.
\vspace{1em} 
\newline
\textbf{Recaption:} \newline
a series of graphs and diagrams illustrating the performance of a rechargeable Zn-CO$_2$ battery using a NiFe-DASC catalyst. Panel (a) shows schematic illustrations of the battery during discharging and charging, depicting the flow of electrons, ions, and gases. Panel (b) presents a chronopotentiometric curve where the voltage is plotted against time, with gas supply indicated as Ar, CO$_2$, and O$_2$. Panel (c) shows a polarization curve with voltage plotted against current density for charge and discharge processes. Panel (d) displays discharge voltage over time at different current densities. Panel (e) plots the Faradaic efficiency of CO and H$_2$ generation as a function of current density. Panel (f) shows galvanostatic discharge-charge cycling curves over 180 cycles. Panel (g) compares the discharge performance of Zn-CO$_2$ batteries using different catalysts, with current density plotted against voltage. \\
\midrule

\includegraphics[width=\linewidth]{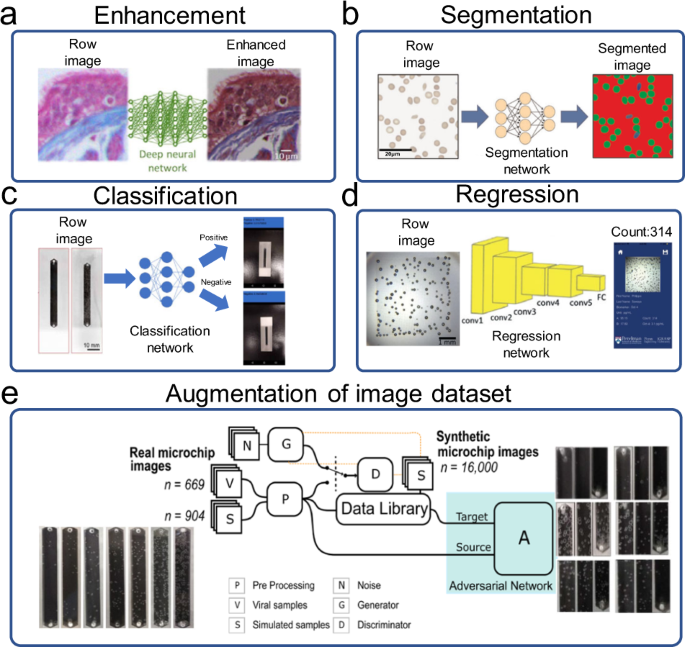} & 
\textbf{Original caption:} \newline 
Fig. 7: Deep learning algorithms for mHealth platforms.
\vspace{1em} 
\newline
\textbf{Recaption:} \newline
a composite figure with five labeled panels (a–e) illustrating different deep learning applications in mobile health platforms. Panel (a) shows an image enhancement process where a raw image is input into a deep neural network to produce an enhanced image. Panel (b) depicts image segmentation, where a raw image is processed by a segmentation network to generate a segmented image. Panel (c) illustrates image classification, where a raw image is fed into a classification network to output a positive or negative result. Panel (d) shows regression, where a raw image is processed by a regression network to produce a numerical count. Panel (e) illustrates the augmentation of an image dataset using a generative adversarial network (GAN), where real microchip images are used to train a generator to produce synthetic microchip images, which are then added to a data library. \\
\midrule

\includegraphics[width=\linewidth]{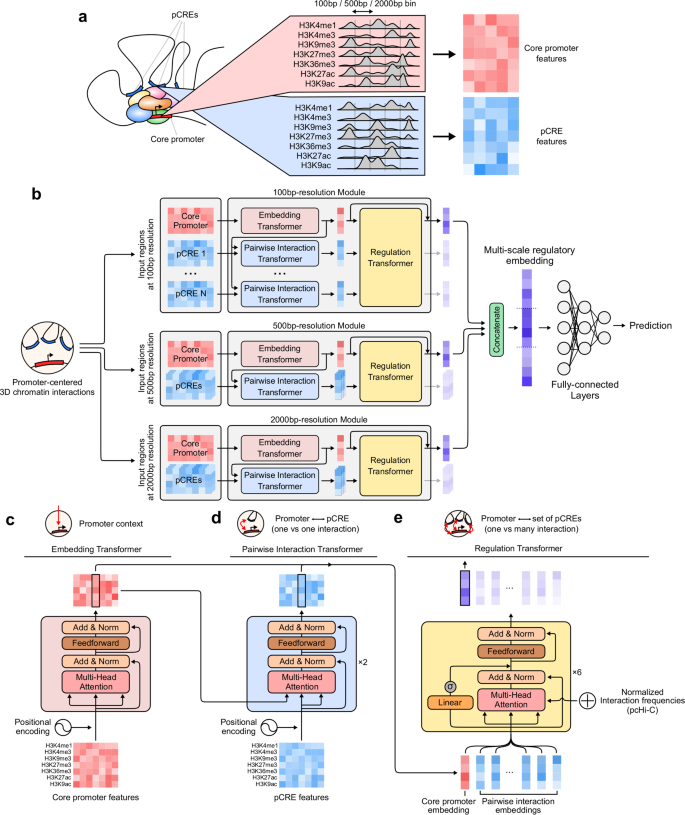} & 
\textbf{Original caption:} \newline 
Fig. 1: Chromoformer model architecture.
\vspace{1em} 
\newline
\textbf{Recaption:} \newline
a schematic of the Chromoformer model architecture, divided into five panels. Panel a illustrates the input features, showing histone modification signals from a core promoter and putative cis-regulatory regions (pCREs) at different genomic resolutions (100bp, 500bp, 2000bp), which are converted into core promoter and pCRE features. Panel b presents the overall model structure, consisting of three parallel modules (100bp-resolution, 500bp-resolution, and 2000bp-resolution) that process input features at their respective resolutions. Each module includes a Core Promoter block, a pCRE block, an Embedding Transformer, a Pairwise Interaction Transformer, and a Regulation Transformer, with outputs from the three modules concatenated to form a multi-scale regulatory embedding. This embedding is then passed through fully-connected layers to generate a prediction. Panels c, d, and e detail the internal architecture of the three transformer components: the Embedding Transformer (c), the Pairwise Interaction Transformer (d), and the Regulation Transformer (e). Each transformer block contains layers with Add \& Norm, Feedforward, and Multi-Head Attention operations, with positional encoding applied. The Regulation Transformer also incorporates normalized interaction frequencies (pChI-C) as an input to its self-attention mechanism. \\
\midrule

\includegraphics[width=\linewidth]{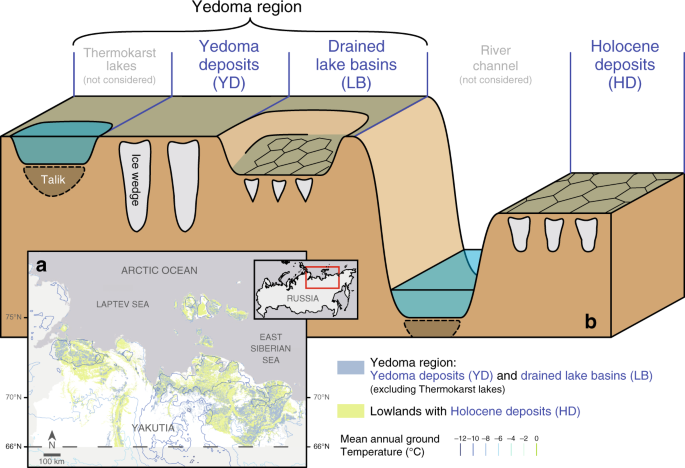} & 
\textbf{Original caption:} \newline 
Fig. 1: Map and schematic of ice-rich permafrost landscapes.
\vspace{1em} 
\newline
\textbf{Recaption:} \newline
a composite scientific illustration consisting of a map and a schematic cross-section of ice-rich permafrost landscapes in northeast Siberia. The map in the lower-left corner shows the geographical extent of the region, including the Arctic Ocean, Laptev Sea, East Siberian Sea, and Yakutia, with a red box indicating the study area. A color-coded legend below the map identifies three main landform types: Yedoma region (Yedoma deposits (YD) and drained lake basins (LB)), lowlands with Holocene deposits (HD), and a temperature scale for mean annual ground temperature in degrees Celsius. The main schematic, labeled 'b', depicts a cross-section of the subsurface, illustrating the stratigraphy and features of these landforms. It shows Yedoma deposits (YD) and drained lake basins (LB) on the left, with a river channel (not considered) and Holocene deposits (HD) on the right. The schematic includes labels for 'Thermokarst lakes (not considered)', 'Talik', 'Ice wedge', and 'Yedoma region'. The diagram illustrates the vertical arrangement of these deposits and features, including the presence of ice wedges within the Yedoma and drained lake basin areas, and the layer of Holocene deposits covering the landscape on the right.\\
\midrule

\end{longtable}

\end{document}